\documentclass[conference]{IEEEtran}
\IEEEoverridecommandlockouts

\usepackage{cite}
\usepackage{amsmath,amssymb,amsfonts}
\usepackage{algorithmic}
\usepackage{graphicx}
\usepackage{textcomp}
\usepackage{xcolor}
\usepackage{subcaption}
\usepackage{hyperref}
\usepackage{enumitem}
\usepackage{placeins}
\usepackage{url}
\usepackage{multirow}
\usepackage{booktabs}  
\usepackage{graphicx}
\def\BibTeX{{\rm B\kern-.05em{\sc i\kern-.025em b}\kern-.08em
    T\kern-.1667em\lower.7ex\hbox{E}\kern-.125emX}}
\begin{document}

\title{\huge Free-Text Evaluation of LLMs for 5G Domain Knowledge and Fault Analysis using LLM-as-Judge}

\author{
\IEEEauthorblockN{Rishiraj Sengupta$^{1,2,*}$\thanks{*now at Google}, Sotiris Chatzimiltis$^{1}$, Mohammad Shojafar$^{1}$, Xiatian Zhu$^{2,3}$}
\IEEEauthorblockA{$^{1}$5G/6G Innovation Centre, Institute for
Communication Systems, University of Surrey, Guildford, Surrey, UK \\
$^{2}$Centre for Vision, Speech and Signal Processing (CVSSP),
University of Surrey, Guildford, Surrey, UK \\
$^{3}$Surrey Institute for People-Centered Artificial Intelligence\\
Email: \{rs02294, sc02449, m.shojafar, xiatian.zhu\}@surrey.ac.uk}
}

\maketitle

\begin{abstract}
Real-world fault analysis in 5G and emerging 6G networks demands domain expertise to analyze free-text diagnostics, including root-cause explanations and recommended actions. LLMs have emerged as a promising approach to automating this, yet whether lightweight, edge-deployable models are capable of performing in-depth free-text diagnostics remains an open question. While existing benchmarks rely on restrictive MCQs with fixed answer keys, this paper evaluates 5G domain understanding and fault analysis in a free-text generation format. Transitioning to this paradigm requires evaluating lightweight, edge-deployable AI models on open-ended diagnostic reasoning, alongside a dependable framework to validate these text outputs at scale. To address this we evaluate three lightweight LLMs, Claude-Haiku-4.5, GPT-5.4-Mini, and Gemini-3.1-Flash-Lite, on free-text 5G domain knowledge and fault-analysis tasks across three benchmarks, TeleQNA\_ORAN\_FT, 5G-Faults\_FT, and TeleInter\_FT. Three independent frontier judges score outputs, and pairwise inter-judge agreement is measured as an empirical test of the LLM-as-Judge methodology. All three models reach at least 90\% accuracy on fault diagnosis, while zero-shot recall of 3GPP and O-RAN specifications remains the critical gap, with all models scoring below 60\%. Mean inter-judge agreement is at least 0.90 across all runs, indicating that multi-judge LLM scoring produces consistent, reproducible grades for open-ended telecom responses. Operationally, Gemini-3.1-Flash-Lite offers the best efficiency trade-off, combining competitive accuracy with the lowest inference cost and latency, making it the most suitable candidate for production telecom deployments.
\end{abstract}


\section{Introduction}

Fault management in 5G and emerging 6G networks is becoming increasingly challenging as network complexity grows. Failures spanning handover procedures, session establishment and core network signalling are difficult to isolate manually, and traditional rule-based approaches do not scale to the diversity of faults in modern deployments. Large language models (LLMs) have recently emerged as a promising approach to automate root cause analysis (RCA), given their ability to reason over natural-language fault descriptions and interpret protocol specifications. ETSI ISG ENI identifies LLMs and generative AI as key technologies for advancing network operations toward higher autonomy, including cross-layer fault analysis and operations, administration and maintenance (OAM) decision-making~\cite{etsi_wp64_2024}.

A key theme across these initiatives is the need for small, efficient models. For operators, LLM adoption must be evaluated not only on technical performance but on energy consumption, cost, and latency, which pose significant barriers to deployment at scale~\cite{shahid2025largescaleaitelecomcharting}. Operators cannot realistically run frontier-scale models across distributed network elements due to inference cost, latency, and data-privacy constraints. ETSI describes a phased approach in which lightweight models first handle Q\&A and knowledge tasks before progressing to fault diagnosis and fuller autonomous operation~\cite{etsi_wp64_2024}. This raises the first research question: \textit{Can lightweight, cost-efficient LLMs handle 5G specification knowledge and fault analysis tasks?} Answering it has direct implications for standardization, where specification knowledge is the foundation on which any progression toward higher network autonomy depends.

Evaluating these models, however, presents a challenge of its own. Unlike multiple-choice benchmarks such as TeleQNA~\cite{maatouk2023teleqnabenchmarkdatasetassess}, real fault analysis requires free-form responses that include explanations, diagnostic steps, and recommended actions, which cannot be scored against a fixed answer key. Yet automated evaluation at scale demands a practical alternative to human annotation. A growing body of work addresses this through the LLM-as-Judge paradigm, where a frontier model scores free-text outputs on behalf of human evaluators~\cite{zheng2023NEURIPS}. While promising, this approach remains unvalidated for telecom, and the broader literature raises concerns about systematic judge biases, including positional, length, and self-enhancement bias, as well as self-consistency~\cite{tan2025judgebenchbenchmarkevaluatingllmbased}. When multiple frontier judges are deployed, a further question arises as to whether they converge on the same verdicts. This raises the second research question: \textit{Do frontier AI judges agree with each other when evaluating telecom free-text outputs?}

This paper addresses both questions through a controlled evaluation of three lightweight, edge-deployable LLMs, Claude-Haiku-4.5, GPT-5.4-Mini, and Gemini-3.1-Flash-Lite on free-text 5G specification and fault-analysis tasks drawn from the TeleQNA~\cite{maatouk2023teleqnabenchmarkdatasetassess} and ORAN-Bench-13K~\cite{gajjar_oran_bench} datasets. Rather than relying on a single evaluator, we score each response with three independent frontier judges, GPT-5.5, Gemini-3.1-Pro, and Gemini-3.5-Flash, and measure their agreement as an empirical test of the LLM-as-Judge method itself. Alongside correctness, we report inference cost, latency, and token usage as practical indicators of edge deployment viability.

The contributions of this paper are as follows:
\begin{itemize}
    \item \textbf{A free-text evaluation of lightweight LLMs on telecom tasks}, using a dataset rewritten into open-ended form across three benchmarks (TeleQNA\_ORAN\_FT, 5G-Faults\_FT, and TeleInter\_FT) that requires models to generate explanations, diagnostic steps, and recommended actions rather than pick from fixed options.
    \item \textbf{An operational study} of cost-efficient, edge-deployable models on these tasks, reporting inference cost, latency, and token usage alongside correctness as signs of deployment viability.
    \item \textbf{A multi-judge evaluation framework}, one example of automated free-text scoring, using three independent frontier judges and inter-judge agreement analysis to test how reliable LLM-as-Judge is for telecom.
\end{itemize}


\section{Related Work} \label{sec:rel_work}
The first systematic evaluation of LLM telecom knowledge, TeleQNA~\cite{maatouk2023teleqnabenchmarkdatasetassess}, introduced 10,000 multiple-choice questions (MCQs) from 3GPP standards, publications, and lexicons, showing that LLMs can rival active professionals on knowledge tasks. ORAN-Bench-13K~\cite{gajjar_oran_bench} extended this to Open RAN with 13,952 MCQs from 116 O-RAN specifications, finding current models not yet proficient. Tele-LLMs~\cite{maatouk_2026_telellms} then showed that domain-adapted smaller models can outperform general-purpose frontier models, while flagging MCQ selection bias as a methodological limitation. Most recently, 6G-Bench~\cite{ferrag_6g_bench} contributed 10,000 MCQs across 30 decision-making tasks from 6G standardization. All of these efforts share a reliance on multiple-choice format. No free-text evaluation framework exists for the domain.

Beyond benchmarking, a growing body of work targets operational network tasks. Singh~\cite{singh2021leveraging} established the motivation for AI-assisted troubleshooting, citing cost and energy efficiency as primary drivers. TeleLogs~\cite{sana2025reasoninglanguagemodelsroot} introduced a curated dataset of annotated 5G RCA problems from realistic drive-test scenarios, showing that general LLMs struggle with fault analysis without domain adaptation. The AI-on-RAN framework~\cite{chatzimiltis_2026_xaillm} established the feasibility of LLM inference at the RAN level for anomaly detection, providing operational context for edge deployment. None of these, however, evaluate free-text diagnostic outputs or quantify the operational metrics that determine edge viability.

Our evaluation methodology builds on the LLM-as-Judge paradigm of Zheng et al.~\cite{zheng2023NEURIPS}, who showed frontier LLMs can serve as reliable proxies for human judgment on open-ended responses. Gu et al.~\cite{gu2026survey} survey the paradigm, noting that LLMs offer finer-grained assessment than rule-based metrics where semantic alignment matters, as in free-text telecom answers, where many valid phrasings express the same concept. Jung et al.~\cite{jung_2025_trustorescalate} introduce a trust-or-escalate framework with provable human-agreement guarantees, motivating our use of inter-judge agreement rate as a reliability signal, and the DeepMind taxonomy of rankers, judges, and assistants~\cite{deepmind_rankers} grounds our choice of binary correctness judges over preference rankers. Crucially, none applies LLM-as-Judge to telecom, and no prior study has measured inter-judge agreement on telecom free-text outputs.

\section{Methodology}\label{sec:methodology}

\begin{figure}[!t]
  \centering
  \includegraphics[width=0.75\columnwidth]{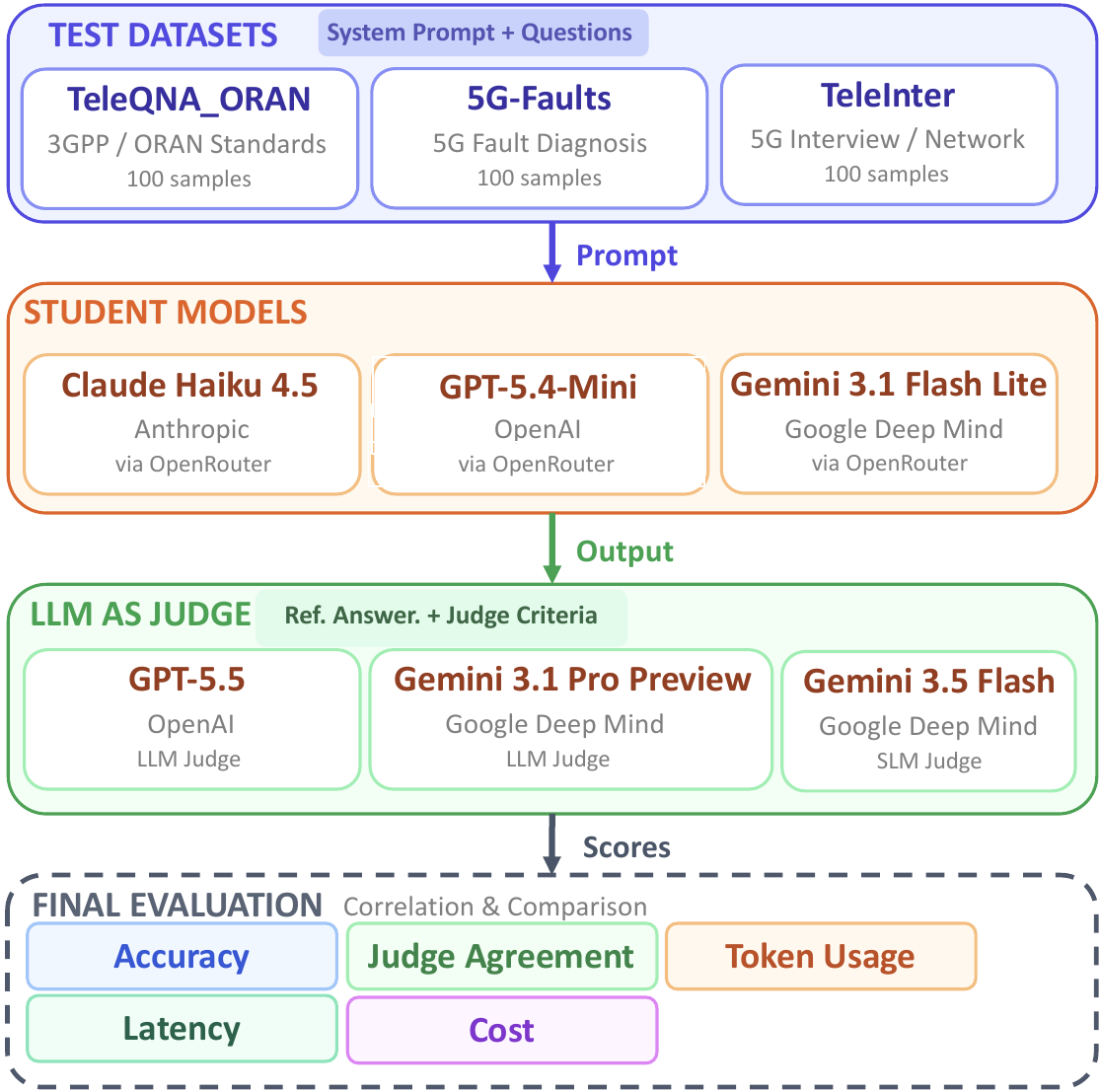}
  \caption{\small Proposed Evaluation Framework. Test datasets are passed
  to student models via the solver pipeline, responses are scored
  by three independent LLM judges, and final metrics covering
  accuracy, judge agreement, token usage, latency and cost are
  computed.}
  \label{fig:framework}
\end{figure}

This section describes the experimental setup used to evaluate lightweight LLMs on telecom free-text tasks. We present the proposed evaluation framework and its components, the student and judge models, the prompting strategy, the benchmark datasets, and the metrics collected. Fig.~\ref{fig:framework} illustrates the end-to-end pipeline.

\subsection{Proposed Evaluation Framework}
The evaluation pipeline is built on Inspect AI~\cite{UK_AI_Security_Institute_Inspect_AI_Framework_2024}, an open-source LLM evaluation framework developed by the UK AI Security Institute. The pipeline extends the LLM evaluation curriculum proposed by Arena Education~\cite{arena_curriculum}, adapting the solver and scorer
components to support expert telecom prompting and multi-judge binary grading. Inspect AI provides deterministic task definitions, reproducible sampling, structured logging in a compressed
\texttt{.eval} format, and a multi-scorer architecture that enables parallel judge invocations within a single evaluation run.

The pipeline is organized into two stages: a solver stage and a scorer stage. The solver stage consists of three sequential components. First, \texttt{prompt\_template()} injects the expert
telecom system prompt and structures the question. Second, \texttt{chain\_of\_thought()} enforces the two-part Student Explanation and Student Answer output template. Third, \texttt{generate()} issues the API call to the student model. The scorer stage invokes all three judge models in parallel, each
independently awarding a binary grade of Correct (C) or Incorrect (I) across the 100-sample run.

All student and judge calls are routed through OpenRouter, a unified gateway over Anthropic, OpenAI, and Google, ensuring consistent rate limiting, token accounting, and cost attribution across providers. Nine evaluation experiments were conducted in total, one per (student model, dataset) combination. For each run, 100 samples are loaded from the corresponding Hugging Face dataset, responses are generated
sequentially through the OpenRouter API with default sampling parameters, and the complete run, including per-sample inputs, outputs, scores and token usage, is persisted as a compressed \texttt{.eval} log file. Accuracy, inter-judge agreement, token usage, latency and cost statistics are then aggregated from the logs.

\subsection{Student Models}
Three lightweight, commercially available LLMs serve as student models in this evaluation. Table~\ref{tab:students} lists each model along with its provider and OpenRouter identifier. The models were selected to represent cost-optimized, low-latency offerings from three major AI providers, namely Anthropic, OpenAI, and Google, making them practical candidates for edge deployment in telecommunications environments. Each model generates a free-text response directly from the prompt without additional inference-time computation.

\begin{table}[!t]
\caption{\small Student Models}
\label{tab:students}
\centering
\begin{tabular}{lll}
\hline
\textbf{Model} & \textbf{OpenRouter ID} & \textbf{Provider} \\
\hline
Claude-Haiku-4.5       & anthropic/claude-haiku-4.5    & Anthropic \\
GPT-5.4-Mini           & openai/gpt-5.4-mini           & OpenAI    \\
Gemini-3.1-Flash-Lite  & google/gemini-3.1-flash-lite  & Google    \\
\hline
\end{tabular}
\end{table}

\subsection{Prompting Strategy}
All student models receive a structured prompt consisting of an expert telecom system prompt followed by the question. To facilitate downstream judge evaluation, each prompt enforces a two-part output template that separates the model's reasoning trace from its final answer. The required response format is as follows:

{\footnotesize
\begin{verbatim}
You are an expert Telecommunication Engineer. 
Provide short, accurate, professional 
and standard-compliant answers.

Your response should be in the following format:
Student Explanation: {explanation in 2 lines}
Student Answer: {final answer in 1 line}
\end{verbatim}
}

This structure enables the judge models to assess factual correctness 
independently of the reasoning trace, ensuring that scoring is based 
on the final answer while retaining the explanation for qualitative 
inspection. Student models operate in a zero-shot setting, without few-shot examples, RAG, or tool use.

\subsection{LLM-as-Judge Evaluation}
Since task responses are open-ended and generative, string-matching and overlap-based metrics such as ROUGE are unsuitable, as there is no fixed ground truth against which to measure similarity. We adopt the LLM-as-Judge paradigm, in which three independent frontier judge models assess each student response against the reference answer and award a binary grade. Table~\ref{tab:judges} lists the judge models used in this evaluation.
Using three independent judges mitigates single-judge bias and enables cross-provider agreement analysis. Each judge receives \textbf{the original question, the student explanation, the student answer, and the reference correct answer}. Each judge is instructed to act as a Telecommunications Domain Expert with 3GPP, 5G, LTE, and O-RAN expertise, evaluating factual correctness primarily from domain knowledge and using the reference answer only as a semantic guide. Each judge outputs a binary grade, either Correct (C) or Incorrect (I), alongside a one-sentence justification. The structured output format is enforced via regex parsing, where grade C maps to a score of 1.0 and grade I to 0.0. Per-judge accuracy is then computed as the mean score across all 100 samples.

\begin{table}[!t]
\caption{\small Judge Models}
\label{tab:judges}
\centering
\begin{tabular}{lll}
\hline
\textbf{Model} & \textbf{OpenRouter ID} & \textbf{Provider} \\
\hline
GPT-5.5                & openai/gpt-5.5                & OpenAI \\
Gemini-3.1-Pro-Preview & google/gemini-3.1-pro-preview & Google \\
Gemini-3.5-Flash       & google/gemini-3.5-flash       & Google \\
\hline
\end{tabular}
\end{table}

\subsection{Datasets}
Three telecom-specific benchmarks are used for the performance evaluation, each comprising 100 questions sampled from curated datasets hosted on the Hugging Face Hub. All multiple-choice options are removed from the original datasets to enforce free-text generation, and the reference answer is passed to the judges as a semantic guide rather than an exact-match target.

\textbf{TeleQNA\_ORAN\_FT}~\cite{maatouk2023teleqnabenchmarkdatasetassess, gajjar_oran_bench} draws from 3GPP 5G and O-RAN specifications, probing factual knowledge of standards, protocols, interfaces and architectural components. Each question requires a concise, technically precise natural language answer, making it the most demanding benchmark in terms of specification recall. 

\textbf{5G-Faults\_FT}~\cite{greenwich157_5gfaults_2024}
presents realistic 5G network fault scenarios derived from operational logs. Models are required to identify fault causes, affected components and recommended remediation actions, evaluating applied diagnostic reasoning in a network operations context.

\textbf{TeleInter\_FT} tests the ability to interpret and explain 5G network concepts, configurations and technical details at the depth expected from experienced network engineers, covering architecture, physical layer, layer-2, layer-3 and advanced features. This benchmark is distinctive in that it requires not only factual recall but also assessment of applied technical concepts, as expected from a domain expert in an original equipment manufacturer (OEM) hiring context. 

All datasets are publicly available on the Hugging Face Hub under the \href{https://huggingface.co/rishieee}{rishieee} namespace. Table~\ref{tab:datasets} summarizes the three benchmarks.
\begin{table}[!t]
\caption{\small Evaluation Datasets}
\label{tab:datasets}
\centering
\begin{tabular}{llll}
\hline
\textbf{Name} & \textbf{Domain} & \textbf{Task} & \textbf{n} \\
\hline
TeleQNA\_ORAN\_FT   & 3GPP / O-RAN specs & Knowledge Q\&A & 100 \\
5G-Faults\_FT  & 5G fault diagnosis & Fault analysis & 100 \\
TeleInter\_FT & 5G / OEM concepts  & Technical Q\&A & 100 \\
\hline
\end{tabular}
\end{table}

\subsection{Metrics}
The following metrics are collected and reported for each (student model, dataset) pair across all nine evaluation runs.
\textbf{Accuracy.} Let $s_{i,j} \in \{0, 1\}$ denote the score
assigned to sample $i$ by judge $j$, where:
\begin{equation}
    s_{i,j} = \begin{cases} 1 & \text{if grade is Correct} \\
    0 & \text{otherwise} \end{cases}
    \label{eq:grade}
\end{equation}
Per-judge accuracy and mean accuracy across all judges are defined as:
\begin{equation}
    \text{Acc}_j = \frac{1}{N} \sum_{i=1}^{N} s_{i,j}
    \label{eq:acc}
\end{equation}

\begin{equation}
    \overline{\text{Acc}} = \frac{1}{J} \sum_{j=1}^{J} \text{Acc}_j
    \label{eq:acc_mean}
\end{equation}
where $N = 100$ is the number of samples per run and $J = 3$ is the number of judge models.

\textbf{Inter-Judge Agreement.} Inspired by the pairwise evaluation
methodology of Zheng et al.~\cite{zheng2023NEURIPS}, let
$c_{i,a,b} \in \{0, 1\}$ denote whether judges $a$ and $b$ agree
on sample $i$, where:
\begin{equation}
    c_{i,a,b} = \begin{cases} 1 & \text{if } s_{i,a} = s_{i,b} \\
    0 & \text{otherwise} \end{cases}
    \label{eq:agree_var}
\end{equation}
Pairwise agreement between judges $a$ and $b$ is then:
\begin{equation}
    P(a = b) = \frac{1}{N} \sum_{i=1}^{N} c_{i,a,b}
    \label{eq:agreement}
\end{equation}
Three pairwise values are computed per run, covering GPT-5.5 vs. Gemini-3.1-Pro-Preview, GPT-5.5 vs. Gemini-3.5-Flash, and Gemini-3.1-Pro-Preview vs. Gemini-3.5-Flash.

\textbf{Token Usage.} Total reasoning and completion tokens  are computed as:
\begin{equation}
    T_{k} = \sum_{i=1}^{N} t_{i,k},
    \quad k \in \{\text{reasoning},\ \text{completion}\}
    \label{eq:tokens}
\end{equation}
where $t_{i,k}$ is the token count of type $k$ for sample $i$.

\textbf{Generation Latency.} Average response latency per sample is defined as:
\begin{equation}
    \overline{\text{Lat}} = \frac{1}{N} \sum_{i=1}^{N} \Delta t_i
    \label{eq:latency}
\end{equation}

where $\Delta t_i$ is the API call duration for sample $i$ in seconds. 

\textbf{API Cost.} Total inference cost per (student model, dataset)
pair is reported in USD and extracted from OpenRouter activity logs.
Cost attribution is matched by model identifier and API key
corresponding to each benchmark task.


\section{Results and Discussion}\label{sec:res_disc}
This section presents and discusses the evaluation results across the three benchmarks for the three student models, covering accuracy, inter-judge agreement, API cost, latency and token usage.

\subsection{Accuracy}

 \begin{figure*}[!t]
  \centering
  \includegraphics[width=\textwidth]{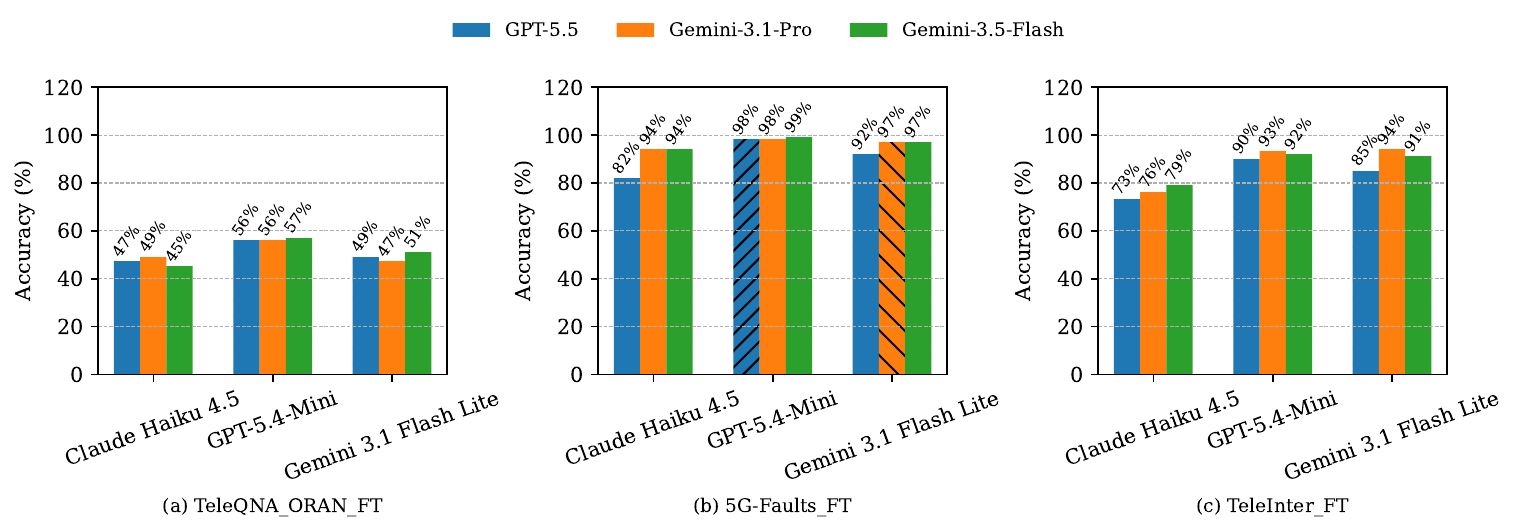}
  \vspace{-2em}
  \caption{\small Accuracy by student model and judge across the three
  benchmarks. Each group of three bars represents one student model,
  coloured by judge model.}
  \label{fig:accuracy}
\end{figure*}

Fig.~\ref{fig:accuracy} reports per-judge accuracy for each (student model, dataset) combination, while Table~\ref{tab:accuracy_mean} summarizes the overall mean accuracy
per model. GPT-5.4-Mini achieves the highest overall mean accuracy at 82.1\%, followed by Gemini-3.1-Flash-Lite at 78.1\% and Claude-Haiku-4.5 at 71.0\%.

TeleQNA\_ORAN\_FT is the most challenging benchmark, with all three models scoring below 60\%, indicating that precise zero-shot recall of 3GPP and O-RAN standards remains difficult and suggesting that retrieval augmentation or domain fine-tuning is needed before deployment. Judge agreement stays high despite the low scores, confirming these figures reflect genuine model limitations rather than evaluator disagreement. 5G-Faults\_FT yields the highest scores overall, with GPT-5.4-Mini reaching near-perfect accuracy at 98.3\%, suggesting 5G fault scenarios are more accessible to lightweight models than open-ended specification knowledge. TeleInter\_FT is intermediate, with GPT-5.4-Mini and Gemini-3.1-Flash-Lite maintaining strong performance while Claude-Haiku-4.5 drops to 76.0\%, a 15 percentage-point gap behind the leader.
 
\subsection{Inter-Judge Agreement}
Table~\ref{tab:agreement} reports the pairwise inter-judge agreement. The agreement rate is the probability that two distinct judges assign the same binary grade to a randomly selected question. Agreement is consistently high throughout, with the two Gemini judges reaching 0.95--0.99 in every
case, reflecting shared provider training and evaluation philosophy. TeleQNA\_ORAN\_FT shows the most balanced cross-provider agreement (0.930--0.970), validating that the low accuracy figures on this benchmark reflect genuine model limitation rather than evaluator disagreement.

The most informative divergence occurs on 5G-Faults\_FT with Claude-Haiku-4.5, where both GPT-5.5 pairwise values drop to 0.860 while
P(Gemini-3.1-Pro = Gemini-3.5-Flash) = 0.980. This pattern reveals that GPT-5.5 systematically disagrees with both Gemini judges on a subset of questions, grading Claude-Haiku-4.5 responses as Incorrect while both Gemini models accept them as Correct. This explains why GPT-5.5 scores Claude-Haiku-4.5 at 82\% while both Gemini judges score it at approximately 94\%, as visible in Fig.~\ref{fig:accuracy}, and points to a provider-level grading style difference on fault diagnosis responses. This finding illustrates the importance of multi-judge evaluation: a single judge would have obscured this systematic divergence entirely.

\begin{table}[!t]
\caption{\small Mean Accuracy per Student Model (\%)}
\label{tab:accuracy_mean}
\centering
\resizebox{\columnwidth}{!}{%
\begin{tabular}{lcccc}
\hline
\textbf{Model} & \textbf{TeleQNA\_ORAN\_FT} & \textbf{5G-Faults\_FT}
& \textbf{TeleInter\_FT} & \textbf{Overall} \\
\hline
Claude-Haiku-4.5       & 47.0 & 90.0 & 76.0 & 71.0 \\
GPT-5.4-Mini           & 56.3 & 98.3 & 91.7 & 82.1 \\
Gemini-3.1-Flash-Lite  & 49.0 & 95.3 & 90.0 & 78.1 \\
\hline
\end{tabular}}
\end{table}

\begin{table}[!t]
\caption{\small Pairwise Inter-Judge Agreement Rates}
\label{tab:agreement}
\centering
\small
\setlength{\tabcolsep}{4pt} 

\begin{tabular}{lcccc}
\toprule
\textbf{Model} &
\textbf{GPT/Pro} &
\textbf{GPT/Flash} &
\textbf{Pro/Flash} &
\textbf{Mean A.R.} \\
\midrule
\multicolumn{5}{c}{\textit{TeleQNA\_ORAN\_FT}} \\
\midrule
Haiku & 0.960 & 0.940 & 0.960 & 0.953 \\
Mini  & 0.940 & 0.930 & 0.970 & 0.947 \\
Flash & 0.960 & 0.940 & 0.960 & 0.953 \\
\midrule
\multicolumn{5}{c}{\textit{5G-Faults\_FT}} \\
\midrule
Haiku & 0.860 & 0.860 & 0.980 & 0.900 \\
Mini  & 0.980 & 0.970 & 0.990 & 0.980 \\
Flash & 0.930 & 0.950 & 0.980 & 0.953 \\
\midrule
\multicolumn{5}{c}{\textit{TeleInter\_FT}} \\
\midrule
Haiku & 0.930 & 0.920 & 0.950 & 0.933 \\
Mini  & 0.950 & 0.960 & 0.990 & 0.967 \\
Flash & 0.890 & 0.920 & 0.950 & 0.920 \\
\bottomrule
\end{tabular}

\vspace{1mm}
\footnotesize
\parbox{\columnwidth}{%
\textit{Note:} GPT = GPT-5.5, G-Pro = Gemini-3.1-Pro-Preview, G-Flash = Gemini-3.5-Flash, Haiku = Claude-Haiku-4.5, Mini = GPT-5.4-Mini, Flash = Gemini-3.1-Flash-Lite, Mean A.R. = Mean Agreement Rate.
}
\vspace{-2em}
\end{table}

\subsection{API Cost}

\begin{figure}[!t]
 \vspace{-1em}
  \centering
  \includegraphics[width=\columnwidth]{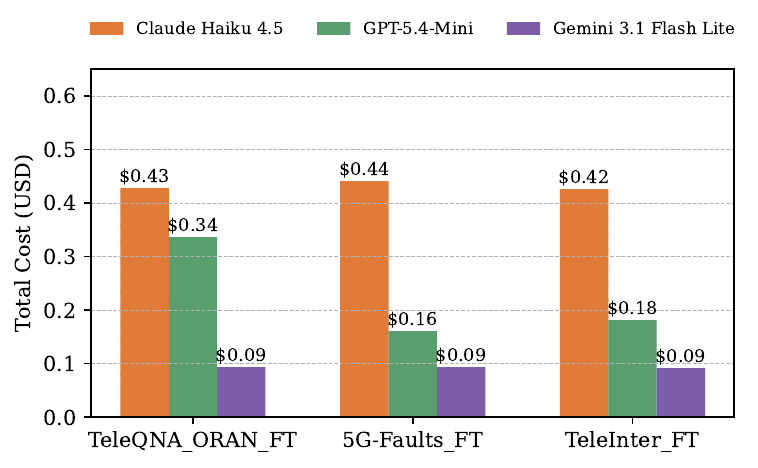}
  \caption{\small Total API inference cost per student model across the
  three benchmarks.}
  \label{fig:cost}
  \vspace{-1.5em}
\end{figure}

Fig.~\ref{fig:cost} reports the total API inference cost per student model across all three benchmarks. Gemini-3.1-Flash-Lite is consistently the most cost-efficient model, remaining below \$0.10 per 100-sample run across all datasets. Claude-Haiku-4.5 is the most expensive, exceeding \$0.42 in every case, while GPT-5.4-Mini sits between the two. Considering that GPT-5.4-Mini achieves the highest overall accuracy at 82.1\% while Gemini-3.1-Flash-Lite achieves 78.1\% at approximately 4.5x cheaper than Claude-Haiku-4.5 and 2.5x cheaper than GPT-5.4-Mini, Gemini-3.1-Flash-Lite offers the most compelling accuracy-to-cost trade-off for edge deployment in telecommunications environments.

\subsection{Operational Metrics}
Table~\ref{tab:tokens} reports token usage across all nine runs. Claude-Haiku-4.5 consistently generates the highest total token volumes, with reasoning tokens accounting for the majority of its output across all benchmarks, reflecting an extended internal reasoning process regardless of task difficulty. GPT-5.4-Mini shows a different profile, with reasoning tokens dominating only on TeleQNA\_ORAN\_FT, suggesting deeper reasoning is engaged primarily for specification-heavy queries. Gemini-3.1-Flash-Lite produces the lowest completion token volumes across all datasets, indicating a concise generation behavior that contributes directly to its latency and cost advantage.

\begin{figure}[!t]
  \centering
  \includegraphics[width=\columnwidth]{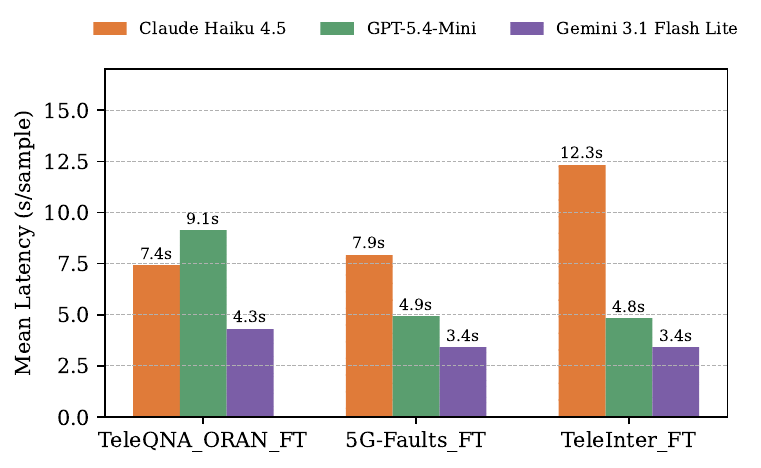}
  \vspace{-2em}
  \caption{\small Mean generation latency per sample (seconds) across the
  three benchmarks.}
  \label{fig:latency}
  \vspace{-1.5em}
\end{figure}

Fig.~\ref{fig:latency} reports mean generation latency per sample. Gemini-3.1-Flash-Lite achieves the lowest latency in every case (3.4s--4.3s), while Claude-Haiku-4.5 peaks at 12.3s on TeleInter\_FT, consistent with its heavy reasoning token usage. GPT-5.4-Mini records its highest latency on TeleQNA\_ORAN\_FT at 9.1s, but improves considerably on the remaining benchmarks. Taken together with Fig.~\ref{fig:cost}, Gemini-3.1-Flash-Lite demonstrates the most favorable operational profile, combining competitive accuracy with the lowest inference cost, latency, and token output across all benchmarks.

\begin{table}[!t]
\caption{\small Token Usage per Student Model}
\label{tab:tokens}
\centering
\small
\setlength{\tabcolsep}{4pt}
\resizebox{\columnwidth}{!}{%
\begin{tabular}{llccc}
\toprule
\textbf{Dataset} & \textbf{Student Model}
& \textbf{Completion} & \textbf{Reasoning} & \textbf{Total} \\
\midrule
\multirow{3}{*}{TeleQNA\_ORAN\_FT}
 & Haiku & 25,704 & 57,033 & 82,737 \\
 & Mini  &  8,236 & 65,002 & 73,238 \\
 & Flash &  8,215 & 51,452 & 59,667 \\
\midrule
\multirow{3}{*}{5G-Faults\_FT}
 & Haiku & 23,919 & 60,625 & 84,544 \\
 & Mini  & 13,623 & 20,116 & 33,739 \\
 & Flash & 11,212 & 48,523 & 59,735 \\
\midrule
\multirow{3}{*}{TeleInter\_FT}
 & Haiku & 27,876 & 54,470 & 82,346 \\
 & Mini  & 11,991 & 26,788 & 38,779 \\
 & Flash &  9,847 & 48,855 & 58,702 \\
\bottomrule
\end{tabular}}

\vspace{0.4mm}
\footnotesize
\parbox{\columnwidth}{%
\textit{Note:} Haiku = Claude-Haiku-4.5, Mini = GPT-5.4-Mini, Flash = Gemini-3.1-Flash-Lite.
}
\vspace{-2.2em}
\end{table}

\section{Conclusions}\label{sec:conclusions}
This paper presented a free-text evaluation of lightweight LLMs on telecom tasks, a more challenging and practically relevant setting than multiple-choice benchmarks. Across three telecom-specific benchmarks, all three models reached at least 90\% accuracy on 5G fault diagnosis, with two of three also reaching 90\% on network interpretation, showing that lightweight models are already capable on structured free-text tasks. Zero-shot recall of 3GPP and O-RAN specifications remains the critical gap, with all models below 60\%, pointing to the need for retrieval-augmented or domain-adapted approaches. The three-judge panel maintained inter-judge agreement of 0.90–0.98 across all nine runs, indicating stable, reproducible grading of open-ended telecom responses. We note this establishes consistency among judges rather than ground-truth correctness, motivating human-expert validation. Among the models evaluated, Gemini-3.1-Flash-Lite offers the best efficiency trade-off, combining competitive accuracy with the lowest cost and latency, making it the strongest candidate for production telecom deployment.

Future work will focus on three directions. First, incorporating human expert judgment alongside the automated judge panel to assess how well LLMs align with domain specialist evaluation in a telecom context, providing external validation of the LLM-as-Judge methodology. Second, addressing the specification knowledge gap identified on TeleQNA\_ORAN\_FT through retrieval-augmented generation and domain fine-tuning, and assessing whether the resulting models meet the standards-recall threshold needed for autonomous OAM operation. Third, expanding from this baseline capability assessment to a more advanced study of Agentic AI frameworks, where multiple specialized models leverage tools and search capabilities to collaborate on complex network tasks.

\section*{Acknowledgment}
This work is supported by the CHEDDAR (Communications Hub for Empowering Distributed Cloud Computing Applications and Research) program, funded by the UK EPSRC under grant numbers EP/Y037421/1 and EP/X040518/1, and in part by the Open Challenge Fund Call through Datacom Industry Association AISBL (the specific call being sponsored by Huawei Technologies).

\bibliographystyle{IEEEtran}
\bibliography{refereces}

\end{document}